\documentclass[letterpaper,10pt,conference]{ieeeconf}
\IEEEoverridecommandlockouts

\usepackage{rotating}
\usepackage{float}
\usepackage{adjustbox}
\usepackage{url}
\usepackage{stfloats}
\usepackage{amsfonts}
\usepackage{multirow}
\usepackage{graphicx}
\usepackage{amsmath}

\usepackage{todonotes}

\title{\LARGE \bf
Low-cost concept-based localized explanations: How far can we get with training-free approaches?}

\author{Darian Fernández-Gutiérrez$^{1, 2}$*, Rafael Bello$^{1}$, Marilyn Bello$^{2}$ and Natalia Díaz-Rodríguez$^{2}$
\thanks{$^{1}$ Dept. of Computer Science, Central University "Marta Abreu" of Las Villas (UCLV), 50100 Santa Clara, Cuba}
\thanks{$^{2}$ Dept. of Computer Science and Artificial Intelligence, University of Granada (UGR), 18071 Granada, Spain.}
\thanks{*Corresponding author: {\tt\small dfgutierrez@correo.ugr.es}
}
}

\begin{document}

\maketitle
\thispagestyle{empty}
\pagestyle{empty}

\begin{abstract}

Concept-based Explainable AI (C-XAI) seeks human-understandable explanations grounded in semantic concepts, yet validation is limited by the scarcity of fine-grained concept annotations. We evaluate whether mid-scale Multimodal Large Language Models (MLLMs) can perform localized concept naming under strict zero-shot conditions by assigning labels to bounding-box regions at both object and part levels. We propose a reproducible zero-shot evaluation protocol for \textbf{Co}ncept \textbf{Na}ming (CoNa) with (i) closed-set, category-constrained prompting for moderate vocabularies and (ii) Open-CoNa, an embedding-similarity-based strategy for large label spaces. Experiments with four MLLMs (7B–32B) show consistent performance trends across datasets, reaching 62\%–88\% object-level exact-match accuracy, highlighting the potential of training-free concept annotation from localized regions. We discuss limitations and failure modes and release a reproducible framework to support future low-cost C-XAI research.

\end{abstract}

\section{Introduction}
The zero-shot capabilities of recent MLLMs have motivated their use in multimodal explainability, where language provides rapid annotation capabilities for images using the closest representation in text (e.g., CLIP \cite{radford_learning_2021}).
However, multimodal models typically rely on paired image--text training, while current eXplainable Artificial Intelligence (XAI) \cite{ali_explainable_2023, bello_three-level_2025} methods remain limited in delivering low-level, fine-grained explanations for such systems. Fine-grained annotations are often unavailable, so explanations usually remain coarse and provide limited support for detailed inspection. Moreover, even if the decision process is transparent, this does not guarantee auditability, particularly for non-technical audiences \cite{arrieta_explainable_2019}. 

Concept-based XAI (C-XAI) aims to make explanations more human-aligned by grounding them in natural-language semantic concepts rather than raw features (e.g., pixels) \cite{poeta_concept-based_2023}. This aligns with transparency requirements for high-risk AI systems, which must communicate their behavior to diverse stakeholders (UNESCO principles and the EU AI Act) \cite{diaz-rodriguez_connecting_2023}.

The recent success of MLLMs as tools for zero-shot learning (ZSL) makes textual encoding of visual concepts useful for automatic annotation of interpretable visual concepts. However, a challenge in the context of explaining visual language models (VLMs) lies in the model understanding (i.e., identifying and localizing visually) granular concepts\footnote{This distinction is essential in concept-based XAI approaches, where explanations must rely on semantic units that are clear, coherent, and understandable to end users \cite{kim_interpretability_2018}.}, i.e., understanding fine- and coarse-grained level concepts localized in any region of an image. This has been demonstrated by VLMs showing language-derived biases \cite{mendez_cubic_2025}.

Since manual object- and part-level annotation is expensive and difficult to scale \cite{ghorbani_towards_2019}, it is important to evaluate whether current small- and mid-scale  MLLMs can reliably assign semantic labels to localized regions under strict \textit{zero-shot} settings. This capability would support semi-automatic annotation workflows for fine-grained C-XAI. However, the impact of factors such as vocabulary size, region type, prompt design, and hallucinations on performance remains insufficiently understood \cite{li_evaluating_2023}.

The goal of this work is to conduct a rigorous and systematic evaluation of small- and mid-scale  multimodal models (up to 32B parameters) to determine their capability to identify localized visual concepts in \textit{zero-shot} mode. To this end, we propose a controlled experimental protocol in which each model receives an image and a delimited region (bounding box) along with a closed set of candidate labels, from which it must select the most appropriate concept category for the visual content. The analysis is performed at both object- and the fine-grained part-level, enabling a comparison of the models' sensitivity to different degrees of semantic granularity. 
The main contributions of this study are as follows:
\begin{itemize}
    \item A \textit{zero-shot} evaluation protocol, called Concept Naming (CoNa) and Open-CoNa, to measure the ability of small- and mid-scale  MLLMs to assign semantic labels to localized visual regions. This pipeline takes as input an image together with the corresponding bounding boxes to ensure a controlled, region-specific evaluation. 
    \item An analysis of several MLLMs in object- and part-level visual naming accuracy, as well as ``unknown'' category rates, evaluated using the PASCAL-Part, ADE20K, and LIP datasets, all of which contain concept-level annotated ground truth.
    \item Empirical evidence showing the potential and limitations of mid-scale  MLLMs for their future integration into semi-automatic conceptual dataset construction workflows. 
\end{itemize}

\section{Related Work}

An increasing body of research is examining the capabilities and limitations of multimodal models for zero-shot visual reasoning. These models, pretrained on large-scale image–text corpora, integrate visual and linguistic information within a shared embedding space, enabling them to assign semantic labels to image regions without task-specific fine-tuning. 
While global capabilities are widely studied, their behavior under localized concept-level naming constraints remains less explored. This motivates a systematic evaluation of reliable object and part-level naming.


\textit{Multimodal Models and Zero-Shot Visual Reasoning}

Large-scale multimodal pretraining learns shared vision–language embeddings that align images and text within a unified representation space. CLIP showed that contrastive training on image–text pairs yields strong semantic alignment and enables zero-shot recognition without task-specific fine-tuning, laying the groundwork for many subsequent multimodal foundation models \cite{radford_learning_2021}.


The present study focuses on low- and mid-scale multimodal models (up to 32B parameters), which offer a favorable balance between reasoning capability and deployability in real-world environments. The models evaluated in this work—LLaVA~1.6--7B, Gemma~3--12B, Mistral~Small~3.1 and Qwen2.5-VL--32B—were selected for their various multimodal alignment mechanisms and their relevance for region-level reasoning. However, prior evaluations of these models have focused largely on global tasks such as image captioning, OCR, or VQA, without examining their behavior when constrained to identify localized concepts from closed vocabularies, a requirement that is essential for 
C-XAI.

\textit{Fine-grained labeling by concept-level naming}

Existing concept-extraction techniques, such as TCAV~\cite{kim_interpretability_2018}, ACE~\cite{ghorbani_towards_2019}, and Concept Bottleneck Models~\cite{koh_concept_2020}, rely on curated concept datasets and supervised training pipelines. These methods assume that meaningful, human-aligned concepts are already available but do not address the challenge of automatically obtaining concept labels at either the object or part-level.

Recent work on open-vocabulary detection and part segmentation (e.g., GroundingDINO~\cite{liu_grounding_2025}) can localize or segment novel categories, but these models still depend on datasets with explicit vision–language correspondence annotations or on promptable detectors that require specifying which concept to detect. This prevents evaluating their intrinsic ability to understand human-aligned concepts, and their performance typically degrades at the part-level.


In contrast, our work evaluates whether MLLMs can serve as training-free concept labelers, assigning semantic labels to localized regions without additional supervision. This fills a gap by assessing their intrinsic naming capability under strict zero-shot, low-cost conditions with predefined concept taxonomies.

\textit{Localized Recognition and Zero-Shot Constraints}

The introduction of the Segment Anything Model (SAM)~\cite{carion2025sam3segmentconcepts} decoupled localization from categorization, enabling modular pipelines where one component determines where a region is and another determines what it represents. Following this perspective, we assume regions are accurately localized (e.g., via SAM3) and focus on whether MLLMs can assign coherent concept labels under zero-shot constraints.

Importantly, while promptable segmenters (e.g., SAM3) require concept names as input, 
in order to properly audit C-XAI, we should not assume that neither the user nor the model knows in advance which concepts are present in an image. Despite recent progress, systematic evaluations of mid-scale MLLMs ($\leq$32B) for fine-grained localized concept identification remain scarce, particularly regarding: (1) object- and object-part-level labeling under closed-set constraints, (2) labeling using large label spaces (e.g., ADE20K whose label set does not fit context windows), (3) interactions between preprocessing/resolution and conceptual accuracy, and (4) failure modes such as hallucination, ambiguity, and explicit “unknown” responses.

This study addresses these gaps by proposing a reproducible protocol to assess the reliability of multimodal models as zero-shot visual concept labelers over given concept segmentation masks, providing empirical evidence relevant to concept-based XAI \cite{poeta_concept-based_2023} and transparent AI in safety-critical settings.

\section{Methodology}

\begin{figure}[htbp!]
\centering
\includegraphics[width=\linewidth]{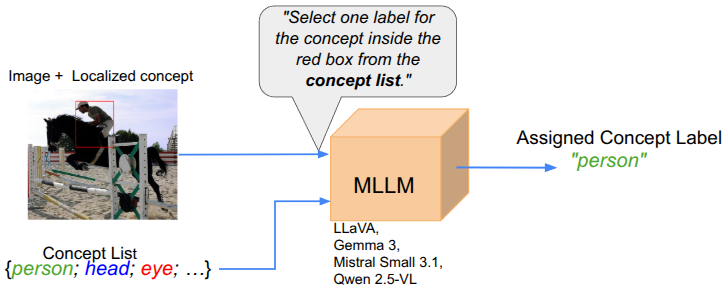}
\caption{Vision Concept Naming (CoNa) pipeline, closed vocabulary setting. 
The input image region and the predefined category dictionary are combined through a concept category-constrained (closed approach) prompt builder, which formulates a structured prompt restricting the MLLM to select exactly one label from the candidate list. The MLLM then produces the most appropriate concept name among the allowed options, yielding the final assigned label.}
\label{fig:CoNa_pipeline}
\end{figure}

This section describes the evaluation protocol designed to measure the ability of low- and mid-scale MLLMs to assign semantic concept labels to localized visual regions under strict zero-shot conditions. The methodology aims to isolate the intrinsic visual–linguistic alignment of each model by standardizing all sources of variability, including category definitions, prompt structure, visual preprocessing, and inference constraints. The overall pipeline is illustrated in Fig.~\ref{fig:CoNa_pipeline}.

\textit{Category Dictionary and Closed-Set Constraints}

For each dataset, a concept dictionary was automatically extracted from the official object and part annotations, ensuring an exact correspondence between ground-truth semantics and the categories available during inference. An additional token, \textit{unknown}, was added to capture ambiguous or unrecognized predictions.

A closed-set regime was adopted for all datasets except ADE20K. Under this setting, models are required to choose exactly one category from the valid vocabulary, preventing vocabulary drift, hallucinated labels, or paraphrastic responses. This constraint enables a cleaner assessment of conceptual alignment by ensuring that any discrepancy between prediction and ground truth reflects semantic inconsistency rather than linguistic variability.

\textit{Prompt Construction and Response Format}:
To maintain consistent textual conditions across models, a standardized, dataset-agnostic prompt builder was implemented so all prompts follow the same template:  

\textit{“Select one label for the object inside the red box. Choose only from: \{...\}. Respond with a single word.”} 

Restricting responses to a single word served three purposes:  
(i) suppressing generative drift and long-form hallucinations common in instruction-tuned models;  
(ii) ensuring comparability across architectures by reducing stylistic variability;  
(iii) allowing an exact-match correspondence between the model outputs and the closed-set dictionary.  
Any output containing multiple tokens or categories outside the dictionary was considered invalid and automatically assigned the \textit{unknown} label. 

\textit{Visual Preprocessing and Region Delimitation}

To ensure uniform visual input, all images were processed through a three-step pipeline: (1) region delimitation using ground-truth bounding boxes or tight crops derived from segmentation masks; (2) contextual cropping with fixed padding to preserve local cues and avoid overly tight crops; and (3) resizing all regions to 512$\times$512 pixels, a resolution selected to balance detail and model capacity. This standardized preprocessing prevents model differences from being driven by crop size, zoom level, or resolution variations.

\textit{Embedding-Based Matching for Large Label Spaces}

\begin{figure*}[htbp!]
    \centering
    \includegraphics[width=\linewidth]{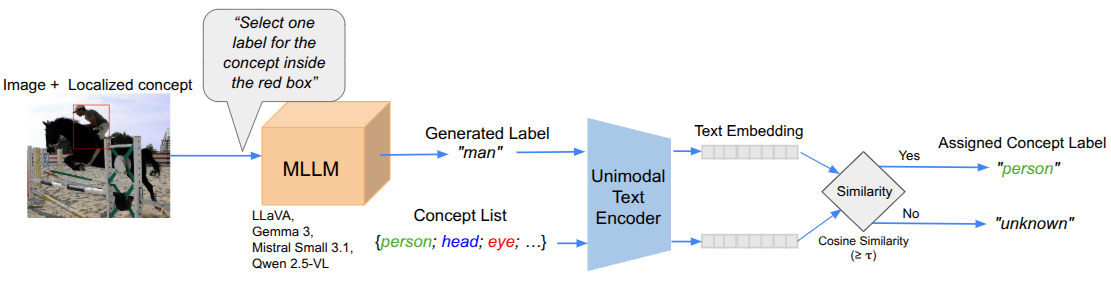}
    \caption{Open-vocabulary Concept Naming (CoNa) pipeline: the MLLM receives no predefined list of possible naming labels. Instead, it generates a free-form single-word description for each input image region. This generated term, together with all categories in the dataset-defined concept dictionary, is encoded into vectors using a text (here nomic-embed-text ~\cite{nussbaum_nomic_2025}) encoder. Cosine similarity ($\geq$ $\tau$) is then computed between the embedding of the generated term and all dictionary embeddings. The dictionary's concept with the highest similarity is selected as the final assigned label, enabling concept labeling without providing all possible labels beforehand.}
    \label{fig:embedding_pipeline}
\end{figure*}

For ADE20K, the closed-set protocol was impractical due to prompt length constraints induced by the large concept dictionary ($\sim$ 3000 labels). In this case, models produced a free-form single-word prediction, which was then mapped to the closest valid category using the embedding-based procedure in Fig.~\ref{fig:embedding_pipeline}.

Both the dictionary labels and the model outputs were encoded using nomic-embed-text~\cite{nussbaum_nomic_2025} text embeddings, because it produces stable embeddings specifically optimized for semantic retrieval and comparison tasks. Cosine similarity was computed between the predicted term and each dictionary entry, assigning the category corresponding to the highest similarity exceeding a threshold $\tau$, which was tuned to maximize agreement with the ground-truth labels. 

Word embeddings were used instead of sentence embeddings for two main reasons:  
(i) the dictionary is taken from the dataset labels and it is primarily composed of decontextualized nouns, and  
(ii) 
for evaluation simplicity (we use 1-gram answers).



\section{Experiments}

\begin{table*}[htbp!]
\caption{Comparison of low- and mid-scale MLLMs evaluated for zero-shot segmented concept naming.
We consider both: Low-cost models (defined as architectures below 10B parameters) and mid-scale models (as those up to 32B parameters). 
\textit{Note:} VRAM refers to the approximate GPU memory required to load and run each model during inference.}
\label{tab:models}
\centering
\renewcommand{\arraystretch}{1.15}
\begin{adjustbox}{max width=\textwidth}
\begin{tabular}{|p{2.4cm}|p{3.7cm}|p{3.1cm}|p{1.8cm}|p{1.8cm}|p{4.2cm}|}
\hline
\textbf{LLM Model Annotator} & \textbf{Base architecture} & \textbf{Main tasks} & \textbf{Parameters} & \textbf{Memory required (VRAM)} & \textbf{Relevance for this study} \\
\hline

LLaVA-1.6--7B~\cite{liu_visual_2023} &
ViT-L/14 + LLaMA-2--7B &
VQA, captioning, basic grounding &
7B (low-scale) &
$\sim$4.7 GB &
Open-source baseline; efficient and widely used for zero-shot spatial grounding. \\

\hline

Gemma 3--12B ~\cite{team_gemma_2025} &
Vision encoder + Gemma Transformer &
Captioning, scene reasoning, OCR &
12B (mid-scale) &
$\sim$8.0 GB &
Lightweight and fast; used to analyze grounding degradation and robustness at moderate scales. \\

\hline

Mistral-Small 3.1 ~\cite{noauthor_mistral_nodate} &
Multimodal encoder + Mistral Transformer &
Captioning, OCR, VQA, dense description &
24B (mid-scale) &
$\sim$15 GB &
Stronger reasoning with moderate cost; good performance–efficiency trade-off in region-level interpretation. \\

\hline

Qwen2.5-VL--32B~\cite{bai_qwen25-vl_2025} &
ViT-L + Qwen2.5 LLM &
Captioning, grounding, visual reasoning &
32B (mid-scale) &
$\sim$21 GB &
Highest-performing open model among those evaluated; top reference for zero-shot semantic labeling of regions. \\

\hline
\end{tabular}
\end{adjustbox}
\end{table*}

\begin{table*}[htbp!]
\caption{Concept-based datasets used to assess the MLLMs' abilities to produce concept-based explanations by asking them to name bounding boxes in images.}
\label{tab:datasets}
\centering
\renewcommand{\arraystretch}{1.1}
\begin{adjustbox}{max width=\textwidth}
\begin{tabular}{|p{2.2cm}|p{2.3cm}|p{2.0cm}|p{2.0cm}|p{2.0cm}|p{2.0cm}|p{6.4cm}|}
\hline
\textbf{Concept-based Dataset} &
\textbf{Number of Images} &
\textbf{\# Object classes} &
\textbf{\# Part classes} &
\textbf{Total Obj.} &
\textbf{Total Parts} &
\textbf{Additional Characteristics} \\
\hline
ADE20K~\cite{zhou_scene_2017} 
& $\sim$10,000 train
& $\sim$1545
& $\sim$330
& 241385
& 127408
& $\sim$3,688 object and stuff categories; provides semantic, instance, and panoptic segmentation for large-scale scene understanding. \\
\hline
PASCAL-Part~\cite{chen_detect_2014} 
& $\sim$10,000 train/val
& 20
& 89
& 24971
& 157464
& 89 fine-grained part categories; contains class- and part-level segmentation masks, enabling compositional part--whole analysis and localized part naming. \\
\hline
LIP (Look-into-Person)~\cite{gong_look_2017} 
& $\sim$10,000 val 
& 1
& 19
& 9972
& 72335
& Pixel-level parsing annotations for human body regions (e.g., arms, legs, torso), optimized for structured human part segmentation. Object-level evaluation corresponds to the person instance, while part labels define the fine-grained taxonomy. \\
\hline
\end{tabular}
\end{adjustbox}
\end{table*}

This section presents the experimental evaluation conducted to analyze the ability of multimodal language–vision models to assign conceptual labels to localized visual regions under strictly zero-shot conditions. We describe the evaluated models, the datasets employed, the initial preprocessing experiment, and the two protocols applied: closed-set CoNa and Open-CoNa.

\textit{Experimental Setup}

The evaluation was designed to quantify the capacity of multimodal models to assign semantically coherent labels under a strictly zero-shot regime. Two levels of visual granularity were considered (objects and parts), ensuring that all models were tested under identical preprocessing, prompting, and inference conditions.

Four representative multimodal models were compared (Table~\ref{tab:models}): LLaVA~1.6 (7B), Gemma~3 (12B), Mistral~Small~3.1 (24B), and Qwen2.5-VL (32B). All were executed in the same containerized environment (Ollama) to guarantee deterministic inference. For each query, the model context was reset, sampling was disabled, and fixed cropping and resolution rules were applied.


Three widely used datasets were included: ADE20K~\cite{zhou_scene_2017}, PASCAL-Part~\cite{chen_detect_2014}, and LIP~\cite{gong_look_2017} (Table~\ref{tab:datasets}). These datasets were selected because they provide a hierarchical concept structure, enabling evaluation with localized regions not only for objects but also for their constituent parts via bounding-box-level supervision (or regions derived from part masks). All annotations were used without modification to preserve closed-set consistency. When a dataset did not include bounding boxes (e.g., ADE20K), we computed them from the dataset ground-truth segmentation masks: for each annotated region, we took the smallest axis-aligned rectangle that fully encloses its mask.

\textit{Initial Preprocessing Experiment on PASCAL-Part }

Before running the main experiments, an initial study was conducted to determine the optimal preprocessing strategy for part-level recognition. PASCAL-Part was selected due to its fine-grained part annotations, which allow precise evaluation of the models' sensitivity to recognize low-level and low-resolution details.

A total of 500 images from the PASCAL-Part dataset were used to compare three configurations: 
(i) tight crops at original resolution;  
(ii) crops resized to 256$\times$256 pixels;  
and (iii) crops resized to 512$\times$512 pixels.

This analysis isolated the effects of resolution and spatial detail on conceptual labeling, using the same prompt and full PASCAL-Part vocabulary. Results showed a consistent improvement with 512$\times$512, particularly for small or highly localized parts. This configuration was therefore adopted for all subsequent experiments in both open and closed CoNa.

\textit{CoNa Protocol: Closed-Set Concept Labeling}

After establishing the optimal preprocessing, the CoNa protocol was applied. Under this setting, the model must select exactly one label from the dataset’s concept dictionary, avoiding linguistic drift and free-form generation. This closed-set regime was applied to all datasets except ADE20K and enabled a direct evaluation of intrinsic visual–semantic alignment.

\textit{Open-CoNa Protocol for ADE20K}

ADE20K contains approximately 3{,}000 categories, making CoNa impractical due to the extreme length of the prompts. For this dataset, the Open-CoNa protocol we propose 
is an embedding-similarity-based concept matching procedure consisting of:  
(1) the MLLM generates a single-word prediction;  
(2) this word and all dataset categories were projected into a unimodal embedding space using the \textit{nomic-embed-text}~\cite{nussbaum_nomic_2025} text encoder;  
(3) cosine similarity is computed between the generated term and all dictionary entries;  
and (4) the final category corresponded to the label with the highest similarity above a predefined threshold\footnote{We performed a random search over the interval $[0.50, 0.75]$ with a step size of $0.05$, obtaining $\tau = 0.60$ as the best-performing threshold.}. 
This approach avoids the closed-set performance degradation caused by a list of potential concept labels that exceeds the model’s context window.

\textit{Evaluation Metrics}

Model performance under open and closed CoNa was quantified using two metrics:  
(i) Exact Match (EM), defined as the proportion of predictions that exactly match the ground-truth label, 
and (ii) Unknown rate (UNK), defined as the proportion of outputs mapped to the \textit{unknown} label, reflecting uncertainty or semantic mismatch. Metrics were computed independently for object classes and object parts.


\section{Results} 

This section reports the empirical findings obtained from the full experimental pipeline. We first describe the outcome of the preliminary evaluation used to determine the optimal preprocessing configuration for part-level analysis, followed by the global results across all models and datasets.

\textit{Preliminary Evaluation on Image Cropping and Resolution Scaling}

A preliminary study was conducted on 500 PASCAL-Part images, comprising 1,231 evaluated objects and 7,847 evaluated parts. The objective was to assess how different preprocessing strategies affect part-level conceptual labeling. Three configurations were compared using Qwen2.5-VL (32B): (i) tight part crops, (ii) crops resized to $256\times256$, and (iii) crops resized to $512\times512$.

Part-level naming performance varies substantially when increasing the crop resolution\footnote{Object-level accuracy (\textit{Obj.~Acc.}) and object-level unknown rates (\textit{UNK Obj.}) remain constant across all configurations because object regions were not modified; only part crops were altered. Accuracy refers exclusively to part-level predictions; object categories were only used to localize regions and were not included in the evaluation.}.
This occurs because low-resolution resizing blurs fine-grained cues (e.g., contours, textures) in small parts, increasing ambiguity. Higher resolutions ($512\times512$) preserve spatial details, reducing the \textit{UNK} rate and improving part-level EM (Table~\ref{tab:preliminary_pascal}). Consequently, we adopted this configuration for all subsequent experiments

\begin{table}[htbp!]
\caption{Ablation study comparing image preprocessing strategies for CoNa using Qwen2.5-VL (32B) on PASCAL-Part. \textit{UNK}nown answers report both percentage and absolute count.
}
\label{tab:preliminary_pascal}
\centering
\renewcommand{\arraystretch}{1.15}
\begin{adjustbox}{max width=\columnwidth}
\begin{tabular}{|p{2.8cm}|c|c|c|c|}
\hline
\textbf{Configuration} &
\textbf{EM Parts $\uparrow$} &
\textbf{UNK Parts $\downarrow$} \\
\hline

Crop-only
& 44.20\%
& 5.29\% (415) \\ \hline

Crop + Resize 256$\times$256
& 46.10\%
& 3.27\% (257) \\ \hline

Crop + Resize 512$\times$512
& \textbf{51.00\%}
& \textbf{3.12\% (245)} \\ \hline

\end{tabular}
\end{adjustbox}
\end{table}

\textit{Full Evaluation Across All Models and Datasets}

Using the $512\times512$ preprocessing configuration—previously identified as the best-performing setting—the evaluation was conducted with CoNa on PASCAL-Part and LIP, where the concept dictionaries define compact and well-structured object and part label spaces. In contrast, Open-CoNa was employed for ADE20K due to its substantially larger and more heterogeneous label set. Table~\ref{tab:full_results} summarizes the global Exact Match (EM) accuracy and Unknown (UNK) rates for both object- and part-level predictions under these configurations.

\begin{table*}[htbp!]
\caption{
CoNa and Open-CoNa (ADE20K) performance across all evaluated MLLMs (with $512\times512$ preprocessing configuration). Exact Match (EM) reports object- and part-level predictions (\%); higher is better ($\uparrow$). Unknown (UNK) and Hallucinations reported as percentage and absolute count; lower is better ($\downarrow$). Hallucination is defined only under closed-set constraints (PASCAL-Part and LIP) as out-of-vocabulary (predictions outside the label set) or format-violating outputs. In ADE20K (Open-CoNa), due to embedding-similarity-based  labeling, hallucinations cannot occur (--).}
\label{tab:full_results}
\centering
\renewcommand{\arraystretch}{1.20}
\begin{adjustbox}{max width=\textwidth}
\begin{tabular}{|l|l|c|c|c|c|c|c|}
\hline
\textbf{Dataset} &
\textbf{Model} &
\textbf{EM Obj. $\uparrow$} &
\textbf{EM Parts $\uparrow$} &
\textbf{UNK Obj. $\downarrow$} &
\textbf{UNK Parts $\downarrow$} &
\textbf{Halluc. Obj. $\downarrow$} &
\textbf{Halluc. Parts $\downarrow$} \\
\hline

\multirow{4}{*}{ADE20K}
& LLaVA 1.6 (7B)          & 4.02\% & 6.13\% & 60.73\% (146583) & 76.9\% (97988)  & -- & -- \\
& Gemma 3 (12B)           & 14.2\% & 21.61\% & \textbf{0.02\% (61)}     & 27.84\% (35473) & -- & -- \\
& Mistral-Small 3.1 (24B) & 19.26\% & 29.04\% & 6.37\% (15384)    & 36.49\% (46491) & -- & -- \\
& Qwen2.5-VL (32B)        &\textbf{ 26.64\% }& \textbf{43.67\%} & 6.58\% (15883)    & \textbf{26.51\% (33780)} & -- & -- \\ \hline

\multirow{4}{*}{PASCAL-Part}
 & LLaVA 1.6 (7B)          & 62.2\%  & 15.5\%  & 10.8\% (2697)     & 28.84\% (45423) & 4.1\% (1016) & 4.45\% (7018) \\
& Gemma 3 (12B)           & 79.2\%  & 37.7\%  & \textbf{0.01\% (4)}        & \textbf{0.35\% (564)}    & 0.5\% (123) & 3.8\% (5989) \\
& Mistral-Small 3.1 (24B) & 81.0\%  & 41.5\%  & 1.6\% (422)       & 3.39\% (5342)   & 0.2\% (51) & 3.77\% (5930) \\
& Qwen2.5-VL (32B)        & \textbf{88.6\%}  & \textbf{51.1\%}  & 0.8\% (201)       & 3.16\% (4984)   & \textbf{0.08\% (19)} & \textbf{2.2\% (3464)} \\ \hline

\multirow{4}{*}{LIP}
& LLaVA 1.6 (7B)          & 81.84\% & 14.41\% & 17.51\% (1746)    & 21.22\% (15346) & 0.05\% (5) & 8.54\% (6180) \\
& Gemma 3 (12B)           & \textbf{99.94\%} & 38.39\% & \textbf{0.0\% (0)}         & \textbf{0.05\% (38)}     & 0.06\% (6) & 0.44\% (319) \\
& Mistral-Small 3.1 (24B) & 98.61\% & 41.45\% & 1.39\% (138)      & 3.30\% (2393)   & \textbf{0.0\% (0)} & 0.26\% (191) \\
& Qwen2.5-VL (32B)        & 97.10\% & \textbf{48.05\%} & 2.5\% (257)       & 3.88\% (2805)   & 0.32\% (32) & \textbf{0.24\% (173)} \\ \hline

\end{tabular}
\end{adjustbox}
\end{table*}

\section{Discussion}

We acknowledge several limitations. Open-CoNa maps free-form outputs via embedding similarity, whereas closed-set prompting offers tighter control over the label space; however, this assumption may not hold in open-world settings with evolving taxonomies. Image resizing can distort very small regions, affecting fine-grained recognition. In addition, using a unimodal text encoder for similarity may not fully capture multimodal semantic geometry and can introduce alignment bias. The single-word constraint reduces linguistic drift but may exclude valid multi-word concepts. Finally, mid-scale MLLMs involve practical trade-offs: inference cost increases with model size, and context-window limits constrain closed-set prompting for large vocabularies such as ADE20K.
Nevertheless, while fine-grained part naming remains challenging, our study shows that training-free approaches can reach strong object-level accuracy and non-trivial part-level performance, positioning mid-scale MLLMs as a cost-effective starting point for concept-based localized explanations.

\section{Conclusion}

This work provides a reproducible zero-shot framework for localized concept labeling using CoNa and Open-CoNa across vocabularies of different sizes. Across three datasets, performance generally improves with model scale, reaching up to nearly 90\% Exact Match at the object level and around 50\% at the part level. Our preprocessing ablation further indicates that $512\times512$ inputs improve part-level naming by preserving spatial detail for small regions. Overall, these results support mid-scale MLLMs as practical, training-free tools for concept annotation in C-XAI, while motivating future work on hierarchical reasoning and evaluation under dynamic or partially known vocabularies.
Code is available\footnote{\url{https://github.com/darianfgUgr/CoNa}}.

\section*{Acknowledgment}
D. Fernández-Gutiérrez gratefully acknowledges the financial support provided by the Junta de Andalucía and the Asociación Universitaria Iberoamericana de Postgrado (AUIP). Díaz acknowledges TSI-100927-2023-1 Project (Transformation and Resilience Plan from the EU NextGen through the Ministry for Digital Transformation and the Civil Service), Google Research Scholar and Grants PID2023-149128NB-I00 and PID2023-150070NB-I00 funded by MICIU/AEI /10.13039/501100011033 and by ERDF, EU, and 
A. Porrello. 
This publication is part of the R\&D\&I project PID2024-156434NB-I00 (CONFIA2), funded by MICIU/AEI/10.13039/501100011033 and FEDER/EU.

\bibliographystyle{IEEEtran}
\bibliography{cai26}

\end{document}